\documentclass[english]{article}
\usepackage{amsfonts}
\usepackage{amsmath}
\usepackage{amssymb}
\usepackage{amsthm}
\usepackage{verbatim}
\usepackage{float}
\usepackage{subfig}
\usepackage[colorlinks,citecolor=blue,urlcolor=blue,linkcolor=red]{hyperref}
\usepackage{fullpage}
\usepackage[affil-it]{authblk}
\usepackage{siunitx}
\usepackage{textcomp}

\usepackage{graphicx}

\renewcommand{\phi}{\varphi}
\renewcommand{\epsilon}{\varepsilon}

\newcommand{\DS}{\mathrm{DS}}
\newcommand{\DC}{\mathrm{DC}}
\newcommand{\MS}{\mathrm{MS}}
\newcommand{\Sp}{\mathrm{Sp}}
\newcommand{\Sn}{\mathrm{Sn}}

\begin{document}
\title{DRUNET: A Dilated-Residual U-Net Deep Learning Network to Digitally Stain Optic Nerve Head Tissues in Optical Coherence Tomography Images}

\author[1]{Sripad Krishna Devalla}
\author[1]{Prajwal K. Renukanand}
\author[1]{Bharathwaj K. Sreedhar}
\author[2,3]{Shamira Perera}
\author[4]{Jean-Martial Mari}
\author[5]{Khai Sing Chin}
\author[1,3]{Tin A. Tun}
\author[3,6,7]{Nicholas G. Strouthidis}
\author[3]{Tin Aung}
\author[5 $\star$]{Alexandre H. Thi\'{e}ry}
\author[1,3 $\star$]{Micha\"el J. A. Girard}

\affil[1]{Ophthalmic Engineering and Innovation Laboratory, Department of Biomedical Engineering, Faculty of Engineering, National University of Singapore, Singapore.}
\affil[2]{Duke-NUS, Graduate Medical School, Singapore.}
\affil[3]{Singapore Eye Research Institute, Singapore National Eye Centre, Singapore.}
\affil[4]{GePaSud, Universit{\'e} de la Polyn{\'e}sie francaise, Tahiti, French Polynesia.}
\affil[5]{Department of Statistics and Applied Probability, National University of Singapore, Singapore.}
\affil[6]{NIHR Biomedical Research Centre at Moorfields Eye Hospital NHS Foundation Trust and UCL Institute of Ophthalmology, London, United Kingdom.}
\affil[7]{Discipline of Clinical Ophthalmology and Eye Health, University of Sydney, Sydney, New South Wales, Australia}
\bigskip
\affil[$\star$]{Both authors contributed equally and are both corresponding authors.}

\maketitle


%
%
\begin{abstract}
Given that the neural and connective tissues of the optic nerve head (ONH) exhibit complex morphological changes with the development and progression of glaucoma, their simultaneous isolation from optical coherence tomography (OCT) images may be of great interest for the clinical diagnosis and management of this pathology. A deep learning algorithm was designed and trained to digitally stain (i.e. highlight) 6 ONH tissue layers by capturing both the local (tissue texture) and contextual information (spatial arrangement of tissues). The overall dice coefficient (mean of all tissues) was $0.91 \pm 0.05$ when assessed against manual segmentations performed by an expert observer. We offer here a robust segmentation framework that could be extended for the automated parametric study of the ONH tissues.
\end{abstract}

%
%
\section{Introduction}
\label{sec.intro}
The development and progression of glaucoma is characterized by complex 3D structural changes within the optic nerve head (ONH) tissues. These include the thinning of the retinal nerve fiber layer (RNFL) \cite{RN29,RN6,RN3}, changes in the minimum-rim-width \cite{RN2}, choroidal thickness \cite{RN1,RN5}, lamina cribrosa (LC) depth \cite{RN30,RN10,RN218}, and posterior scleral thickness \cite{RN11}; and migration of the LC insertion sites \cite{RN32,RN396}. If these parameters (and their changes) could be extracted automatically from optical coherence tomography (OCT) images, this could assist clinicians in their day-to-day management of glaucoma.\\

While there exist several tools to automatically segment the ONH tissues \cite{RN244,RN16,RN517,RN236,RN238,RN243,RN19,RN310,RN255}, and thus extract these parameters, each tissue currently requires a different algorithm (tissue-specific).  In our previous study, we developed a histogram-based \cite{RN239} approach that was able to ‘digitally stain’ (isolate) the connective and neural tissues of the ONH. Following this, we proposed a more accurate deep-learning (patch-based) \cite{RN516} approach to isolate these tissues.\\

Tissue-specific segmentation tools are computationally expensive \cite{RN321} and are often prone to segmentation errors in images with pathologies \cite{RN504}(e.g, age-related macular degeneration, optic disc edema). The segmentation accuracy of these tools is also affected by the reduced deep-tissue visibility and shadow artifacts \cite{RN229} in the OCT images. Combining compensation technology (to remove the deleterious effects of light attenuation) \cite{RN20} with our recently proposed methods, we were able to isolate the ONH tissues more accurately. Yet, the histogram-based approach was limited as it required an initial manual input and could not fully separate different tissues (e.g. sclera from choroid).  On the other hand, the patch-based approach failed to offer precise tissue boundaries, failed to separate the LC from the sclera, and presented artificial LC-scleral insertions.\\

In this study, we present DRUNET (Dilated-Residual U-Net), a novel deep-learning approach capturing both the local (tissue texture) and contextual information (spatial arrangement of the tissues) to digitally stain neural and connective tissues of the ONH. We present a comparison with our earlier deep-learning (patch-based) approach to assert the robustness of DRUNET. Our long-term goal is to offer a framework for the automated extraction of neural and connective tissue structural parameters from OCT images of the ONH.

\section{Methods}
\subsection{Patient Recruitment}
A total of 100 subjects were recruited at the Singapore National Eye Center. All subjects gave written informed consent. This study adhered to the tenets of the Declaration of Helsinki and was approved by the institutional review board of the hospital. The cohort consisted of 40 normal (healthy) controls, 41 subjects with primary open angle glaucoma (POAG) and 19 subjects with primary angle closure glaucoma (PACG). The inclusion criteria for normal controls were: an intraocular pressure (IOP) less than 21 mmHg, healthy optic nerves with a vertical cup-disc ratio (VCDR) less than or equal to 0.5 and normal visual fields test. Primary open angle glaucoma was defined as glaucomatous optic neuropathy (GON; characterized as loss of neuroretinal rim with a VCDR $> 0.7$ and/or focal notching with nerve fiber layer defect attributable to glaucoma and/or asymmetry of VCDR between eyes $> 0.2$) with glaucomatous visual field defects. Primary angle closure glaucoma was defined as the presence of GON with compatible visual field loss, in association with a closed anterior chamber angle and/or peripheral anterior synechiae in at least one eye. A closed anterior chamber angle was defined as the posterior trabecular meshwork not being visible in at least $180$ of anterior chamber angle.

\subsection{Optical Coherence Tomography Imaging}
The subjects were seated and imaged under dark room conditions after dilation with $1\%$ tropicamide solution. The images were acquired by a single operator (TAT), masked to diagnosis with the right ONH being imaged in all the subjects, unless the inclusion criteria were met only in the left eye, in which case the left eye was imaged. A horizontal B-scan (0\textdegree) of 8.9 mm (composed of 768 A-scans) was acquired through the center of the ONH for all the subjects using spectral-domain OCT (Spectralis, Heidelberg Engineering, Heidelberg, Germany). Each OCT image was averaged 48x and enhanced depth imaging (EDI) was used for all scans.

\subsection{Shadow Removal and Light Attenuation: Adaptive Compensation}
We used adaptive compensation (AC) to remove the deleterious effects of light attenuation \cite{RN20}. AC can help mitigate blood vessel shadows and enhance the contrast of OCT images of the ONH \cite{RN20,RN249}. A threshold exponent of 12 (to limit noise over-amplification at high depth) and a contrast exponent of 2 (for improving the overall image contrast) were used for all the B-scans \cite{RN249}.

\subsection{Manual Segmentation}
\begin{figure}[h]
    \centering

    \includegraphics[width=.75\textwidth]{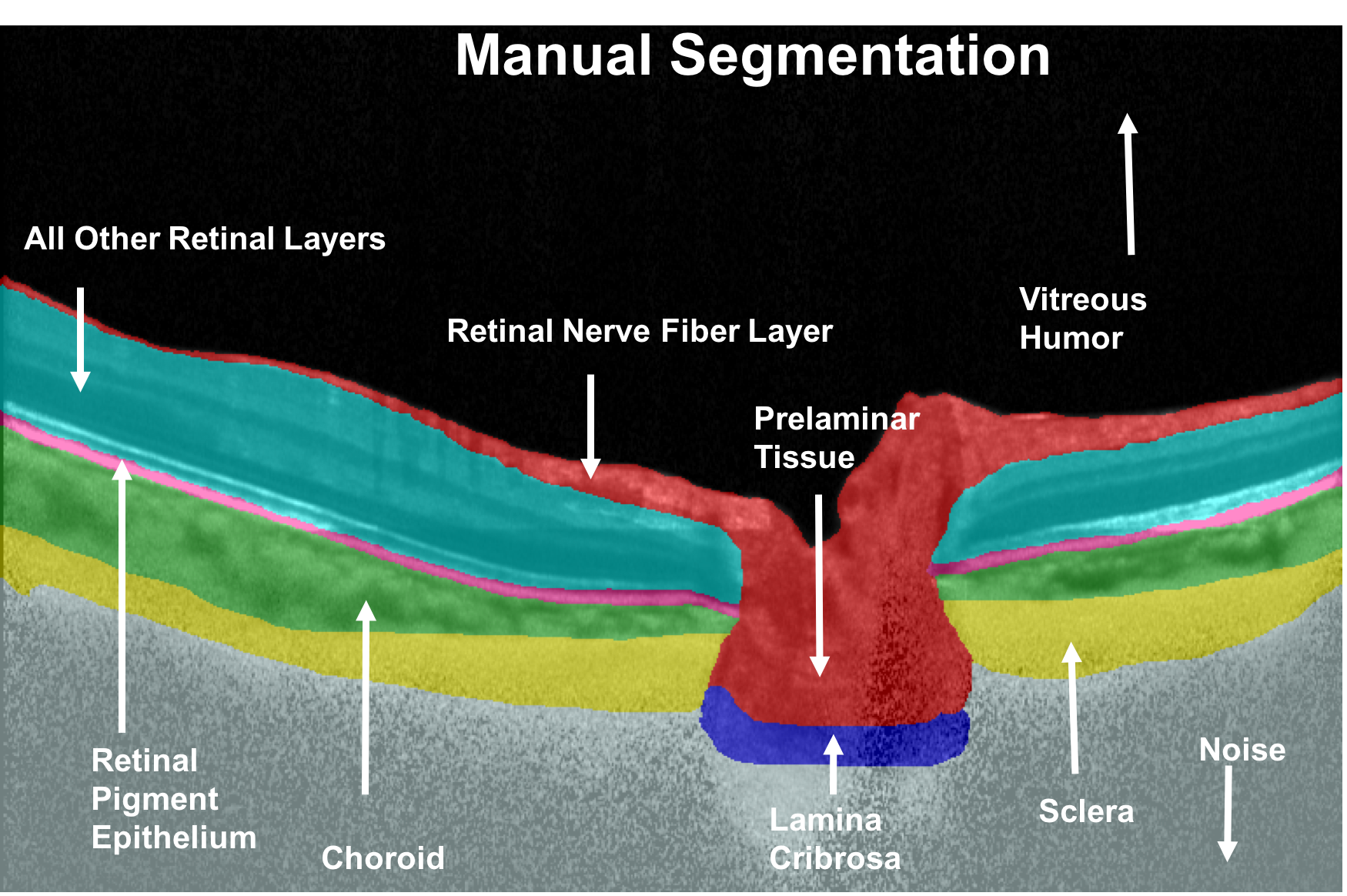}
    \caption{Manual Segmentation of a compensated OCT image. The RNFL and the prelaminar tissue are shown in red, the RPE in pink, all other retinal layers in cyan, the choroid in green, the peripapillary sclera in yellow, the LC in blue, noise in grey and the vitreous humor in black.}
    \label{fig:1}
\end{figure}

An expert observer \textbf{(SD)} performed manual segmentation of all OCT images using Amira (version 5.4, FEI, Hillsboro, OR). This was done to \textbf{1)} train our algorithm to identify and isolate the ONH tissues; and to \textbf{2)} validate the accuracy of digital staining. Each OCT image was segmented into the following classes: (refer \textbf{Figure \ref{fig:1}}):\textbf{(1)} the RNFL and the prelamina (in red); \textbf{(2)} the retinal pigment epithelium (RPE; in pink); \textbf{(3)} all other retinal layers (in cyan); \textbf{(4)} the choroid (in green); \textbf{(5)} the peripapillary sclera (in yellow); and \textbf{(6)} the (LC) (in blue). Noise (in gray) and the vitreous humor (in black) were also isolated.  Note that we were unable to obtain a full thickness segmentation of the peripapillary sclera and the LC due to limited visibility \cite{RN249}. Only their visible portions were segmented.\\

\subsection{Digital Staining of the ONH}
In this study, we developed the architecture \textbf{DRUNET} (Dilated-Residual U-Net): a fully convolutional neural network inspired by the widely used U-Net \cite{RN329}, to digitally stain the ONH tissues. DRUNET exploits the inherent advantages of the U-Net skip connections \cite{RN509}, residual learning \cite{RN23} and dilated convolutions \cite{RN24} to offer a robust digital staining with a minimal number of trainable parameters. The U-Net skip connections allowed capturing both the local and contextual information \cite{RN509,RN329}, while the residual connections allowed a better flow of the gradient information through the network. Using the dilated convolutional filters, we were able to better exploit the contextual information: this was crucial as we believe local information (i.e. tissue texture) is insufficient to delineate precise tissue boundaries. DRUNET was trained with OCT images of the ONH and their corresponding manually segmented ground truths.  \\

\subsection{Network Architecture}
\begin{figure}[H]
    \centering
    \includegraphics[width=1.0\textwidth]{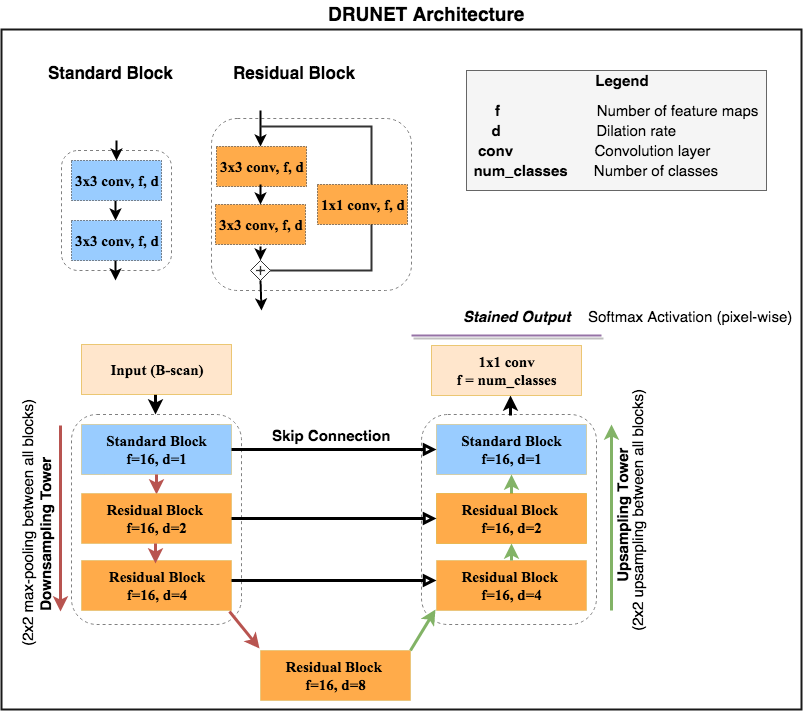}
    \caption{DRUNET comprises of two towers: (1) A downsampling tower – to capture the contextual information (i.e., spatial arrangement of the tissues), and an (2) upsampling tower – to capture the local information (i.e., tissue texture). Each tower consists of two blocks: (1) a standard block, and (2) a residual block. The entire network consists of 40,000 trainable parameters in total. }
    \label{fig:2}
\end{figure}

The DRUNET architecture was composed of a downsampling and an upsampling tower \textbf{Figure \ref{fig:2}}, connected to each other via skip-connections. Each tower consisted of one standard block and two residual blocks. Both the standard and the residual blocks were constructed using two dilated convolution layers, with 16 filters (size 3x3) each. The identity connection in the residual block was implemented using a 1x1 convolution layer, as described in \textbf{Figure \ref{fig:2}}. In the downsampling tower, the input image of size 496x768 was fed to a standard block with a dilation rate of 1 followed by two residual blocks with dilation rates of 2 and 4 respectively. After every block in the downsampling tower, a max-pooling layer of size 2x2 was used to reduce the dimensionality and exploit the contextual information. A residual block with a dilation rate of 8 was used to transfer the features from the downsampling to the upsampling tower. These features were then passed through two residual blocks with dilation rates of 4 and 2 respectively. A standard block with a dilation rate of 1 was used to restore the image to its original resolution. After every block in the upsampling tower, a 2x2 upsampling layer was used to sequentially restore the image to its original resolution. The output layer was implemented as a 1x1 convolution layer with a number of filters equal to the number of classes (8 = 6 tissues + noise + vitreous humour). We then applied a softmax activation to this output layer to obtain the class-wise probabilities for each pixel.  Finally, each pixel was assigned the class of the highest probability. Skip connections \cite{RN509} were established between the downsampling and upsampling towers to recover the spatial information lost during the downsampling.

In both towers, all the layers except the last output layer were batch normalized and activated by an exponential linear unit function ELU \cite{RN27}. In each residual block, the residual layers were batch normalized and ELU activated before their addition.\\

The entire network was trained end-to-end using stochastic gradient descent with nesterov momentum (momentum=0.9). An initial learning rate of 0.1 (halved when the validation loss failed to improve over two consecutive epochs) was used to train the network and the model with the best validation loss was chosen for all the experiments in this study. The loss function $\mathcal{L}$ was based on the mean of Jaccard-type Index calculated for each tissue,

\begin{align*}
\mathcal{L}= 1 - \frac{1}{C} \,  \sum_{c=1}^{C} \frac{\widehat{y}_c \odot y_c}{\widehat{y}_c + y_c - \widehat{y}_c \odot y_c},
\end{align*}
where $C$ denotes the total number of classes, the notation $\widehat{y}_c \odot y_c$ denotes the component-wise multiplication between the matrices $\widehat{y}_c $ and $y_c$, the quantity $\widehat{y}_c \in (0,1)$ is the predicted pixel-wise probability of belonging to class $c \in \{1,2, \ldots, C\}$ and $y_c \in \{0,1\}$ is the ground truth segmentation matrix (or map).
The final network consisted of 40,000 trainable parameters. The proposed architecture was trained and tested on a NVIDIA GTX 1080 founder’s edition GPU with CUDA v8.0 and cuDNN v5.1 acceleration. With the given hardware configuration, each OCT image was digitally stained in 80 ms.

\subsection{Data Augmentation}

An extensive online data augmentation was performed to overcome the sparsity of our training data. Data augmentation consisted of random rotation (8 degrees clockwise and anti-clockwise),  horizontal flipping, nonlinear intensity shifts, addition of white noise and multiplicative speckle noise \cite{RN510}, elastic deformations \cite{RN514} and occluding patches. \\

\begin{figure}[H]
    \centering

    \includegraphics[width=1.\textwidth]{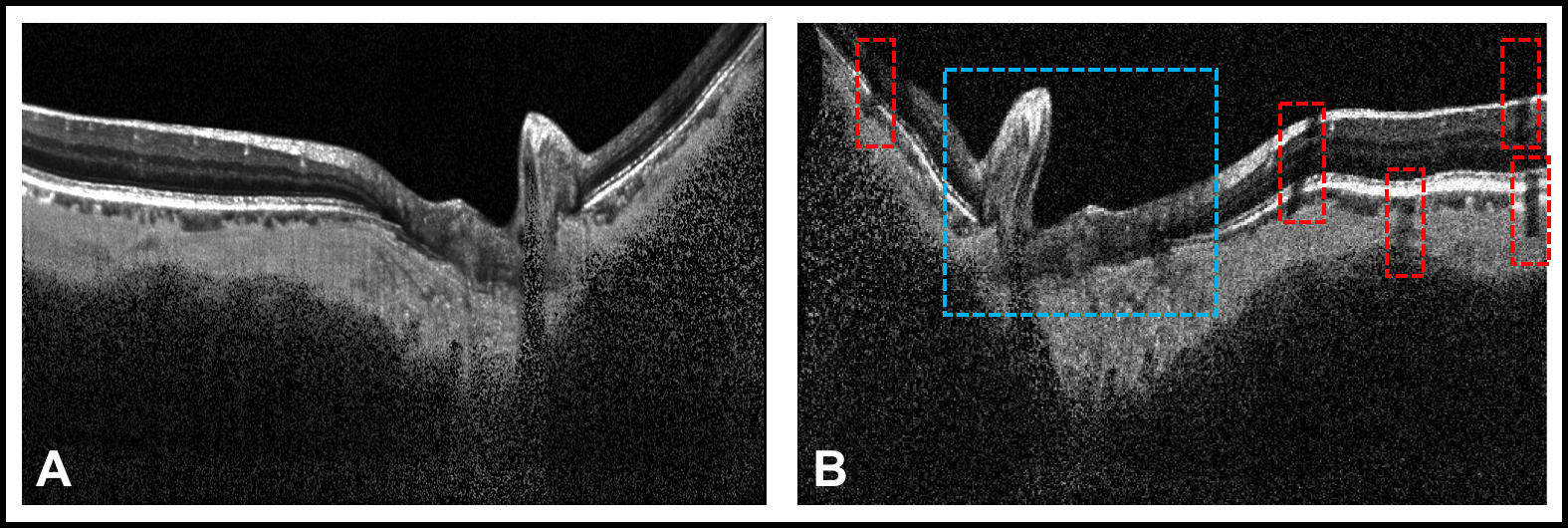}
    \caption{Extensive data augmentation was performed to overcome the sparsity of our training data. \textbf{(A)} represents a compensated OCT image of a glaucoma subject. \textbf{(B)} represents the same image having undergone data augmentation. The data augmentation includes horizontal flipping, rotation (8 degrees clockwise), additive white noise and multiplicative speckle noise \cite{RN510}, elastic deformation \cite{RN514} and occluding patches. A portion of the image undergoing elastic deformation and occlusion from patches is bounded by blue and red box respectively. The elastic deformations (combination of shearing and stretching) made our network invariant to images with atypical morphology (i.e., ONH tissue deformation in glaucoma \cite{RN535}). The occluding patches reduced visibility of certain tissues, making our network invariant to blood vessel shadows.}
    \label{fig:3}
\end{figure}

Nonlinear intensity shifts were of the type $\overline{I} = \phi(I)$, for a random (non-linear) function $\phi$, where the quantities $I$ and $\overline{I}$ denote intensities (pixel-wise) before and after the nonlinear intensity shift respectively. This made the network invariant to intensity inhomogeneity within/between tissue layers (a common problem in OCT images affecting the performance of automated segmentation tools \cite{RN533}). The elastic deformations \cite{RN514} can be viewed as an image warping technique to produce the combined effects of shearing and stretching. This was done in an attempt to make our network invariant to images with atypical morphology (i.e., ONH tissue deformations as seen in glaucoma \cite{RN535}). 
Twenty occluding patches of size 60x20 pixels were also added at random locations to reduce the visibility in certain tissues, in an effort to make our network invariant to blood vessel shadowing that is common in OCT images. Each occluding patch resulted in the reduction of intensity in the entire occluded region by a random factor (random number between 0.2 and 0.8). An example of data augmentation performed on a single OCT image is shown in \textbf{Figure \ref{fig:3}}.

\subsection{Training and Testing of our network}

The dataset of 100 B-scans (40 healthy, 60 glaucoma) was split into training and testing datasets. The training set was composed of an equal number of compensated glaucoma and healthy OCT images of the ONH, along with their corresponding manual segmentations. The trained network was then evaluated on the unseen testing set (composed of the remaining compensated OCT images of the ONH and their corresponding manual segmentations). A training set of 40 images (60 testing images) were chosen for all the experiments discussed in this study.\\

To assess the consistency of the proposed methodology, the model was trained on five training sets of 40 images each and tested on their corresponding testing sets. Given the limitation of a total of only 100 OCT images, it was not possible to obtain five distinct training sets, thus each training set had some images repeated. To study the effect of compensation on digital staining, the entire process (training and testing) was repeated with the baseline (uncompensated) images. A comparative study was also performed between the DRUNET architecture and our previously published patch-based digital staining approach \cite{RN516}. For this, we trained and tested both the techniques with the same dataset.


\subsection{ Digital Staining: Qualitative Analysis}

All the digitally stained images obtained were manually reviewed by an expert observer (SD) and qualitatively compared with their corresponding manual segmentations.\\

\subsection{Digital Staining: Quantitative Analysis}
We used the following metrics to assess the accuracy of the digital staining: (1) the dice coefficient ($\DC$); (2) Specificity ($\Sp$); and (3) ($\Sn$). For each image, the metrics were computed for the following classes: (1) RNFL and prelamina, (2) RPE, (3) all other retinal layers, and (4) choroid. Note that the metrics could not be applied directly to the peripapillary sclera and the LC as their true thickness could not be obtained from the manual segmentation. However, digital staining of the peripapillary sclera and of the LC was qualitatively assessed. Noise and vitreous humor were also exempted from such a quantitative analysis.\\

The dice coefficient was used to measure the spatial overlap between the manual segmentation and the digital staining. It is defined between 0 and 1, where 0 represents no overlap and 1 represents a complete overlap. For each image in the testing set, the dice coefficient was calculated for each tissue as follows,
\begin{align*} 
\DC_i =  2 \, \times \, \frac{\left| \DS_i \cap \MS_i \right|}{ \left| \DS_i \right| + \left| \MS_i\right|}
\end{align*}
where $\MS_i$ is the set of pixels representing the tissue $i$ in the manual segmentation, while $\DS_i$ represents the same in the digitally stained image.\\
Specificity was used to assess the true negative rate of the proposed method.\\
\begin{align*} 
\Sp_i =  \frac{\left| \overline{\DS}_i \cap \overline{\MS}_i \right|}{\left| \overline{\MS}_i\right|}
\end{align*}
where  $\overline{\DS}_i$ and $\overline{\MS}_i$ are the set of all the pixels not belonging to class $i$ in the digitally stained and the corresponding manually segmented image respectively. Sensitivity was used to assess the true positive rate of the proposed method as is defined as:
\begin{align*} 
\Sn_i =  \frac{\left| \DS_i \cap \MS_i \right|}{\left| \MS_i\right|}
\end{align*}
Both specificity and sensitivity were reported on a scale of 0--1. To assess the performance of the digital staining between glaucoma and healthy OCT images, for each experiment, the metrics were calculated separately for the two groups.

%
%
%
%
%
%
%
%

%
%
%
\section{Results}
\subsection{Qualitative Analysis}
\begin{figure}[H]
    \centering
    \includegraphics[width=1.0\textwidth]{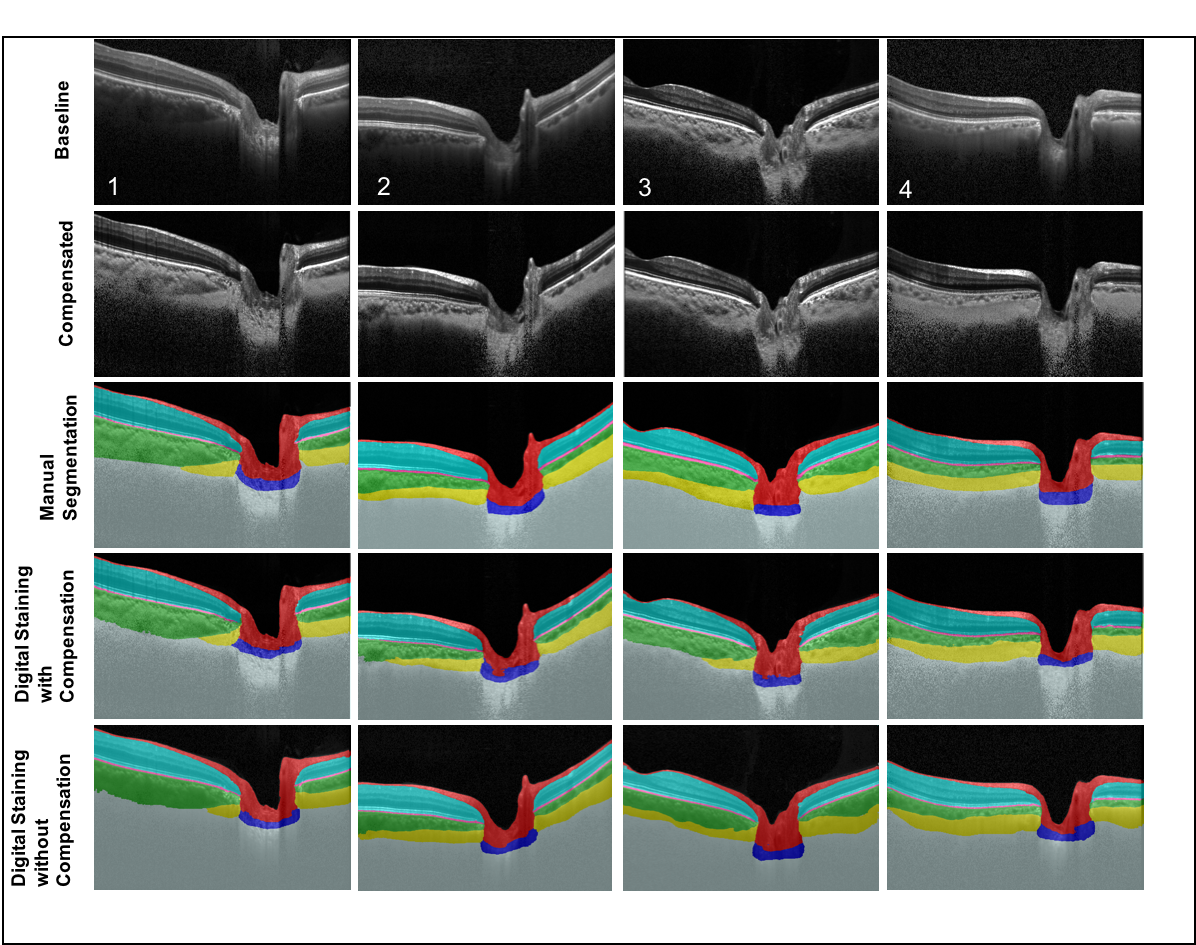}
    \caption{Baseline (1st row), compensated (2nd row), manually segmented (3rd row), digitally-stained images (trained on 10 compensated images; 4th row), and digitally-stained images (trained on 10 baseline images; 5th row) for 4 selected subjects (1$ \& $2: POAG, 3: Healthy, 4:PACG )}
    \label{fig:4}
\end{figure}

The baseline, compensated, manually segmented, and the digitally-stained images for 4 selected subjects (1$ \& $2: POAG, 3: Healthy, 4: PACG) are shown in \textbf{Figure \ref{fig:4}}.\\

When trained with the compensated images (\textbf{Figure \ref{fig:4}, 4th Row}) or the uncompensated images (\textbf{Figure \ref{fig:4}, 5th Row}), DRUNET was able to simultaneously isolate the different ONH tissues, i.e. the RNFL $+$ prelamina (in red), the RPE (in pink), all other retinal layers (in cyan), the choroid (in green), the sclera (in yellow) and the LC (in blue). Noise and vitreous humor were isolated in gray and black respectively. In both cases, the digital staining of the ONH tissues were qualitatively similar, comparable and consistent with the manual segmentation. A smooth delineation of the choroid-sclera interface was obtained in both cases.
Irregular (\textbf{Figure \ref{fig:4}, Subject 2 and 4}) LC boundaries that were inconsistent with the manual segmentations were obtained in few images irrespective of the training data (compensated/uncompensated images). 
When validated against the respective manual segmentations, there was no visual difference in the performance of digital staining on healthy or glaucoma OCT images across all experiments.\\

\subsection{Quantitative Analysis}

\begin{figure}[H]
    \centering

    \includegraphics[width=1.0\textwidth]{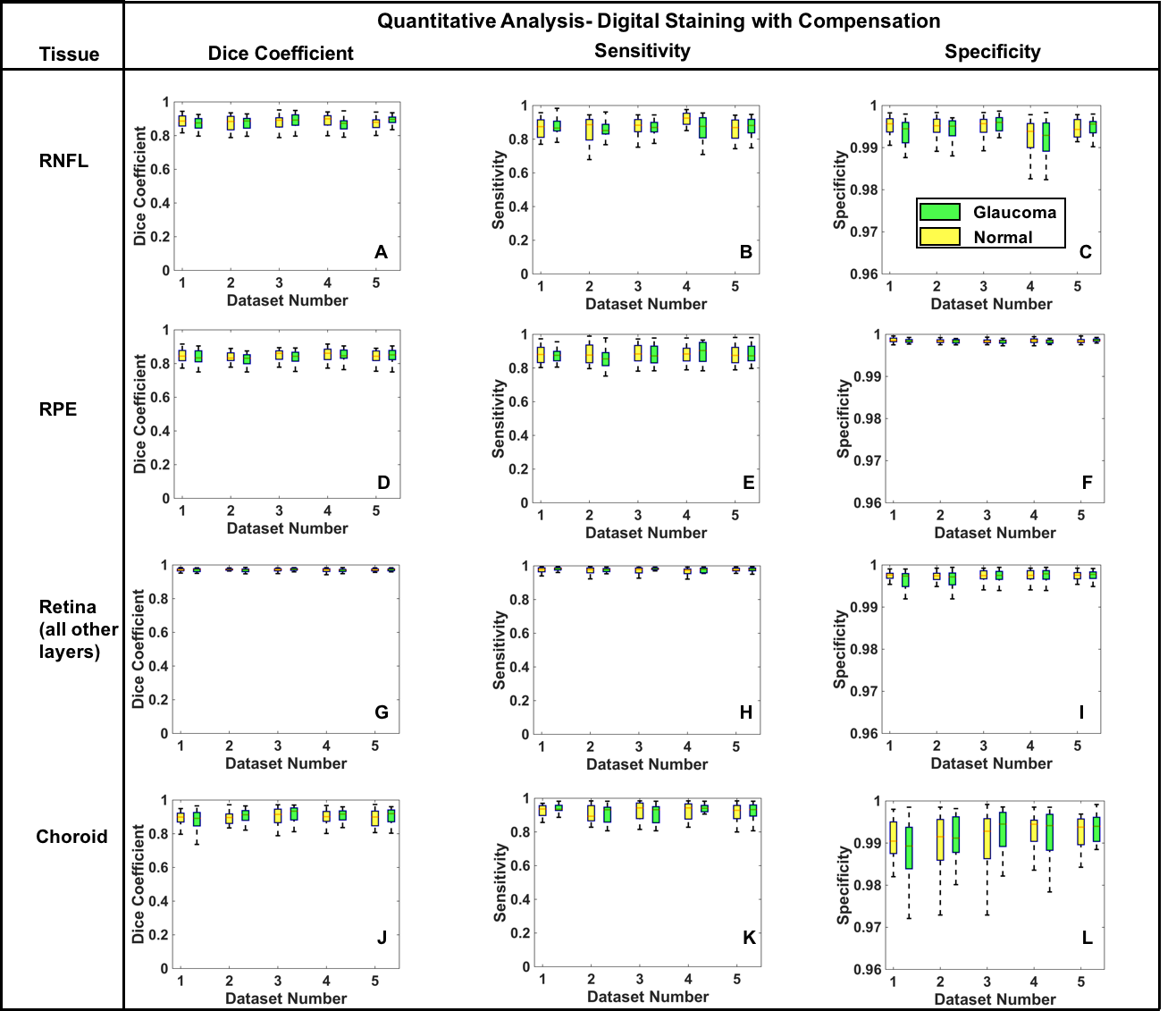}
    \caption{A quantitative analysis of the proposed method is presented to assess the consistency and performance of digital staining between glaucoma and healthy images. A total of 5 datasets were used for training (40 images) and its corresponding testing (60 images). (A-C) represent the dice coefficients, sensitivities and specificities as box plots for the RNFL + prelamina for healthy (in yellow) and glaucoma (in green) images in the testing sets. (D-F) represent the same for the RPE, (G-I) represent the same for all other retinal layers and (J-L) represent the same for the choroid.}
    \label{fig:5}
\end{figure}

When trained with compensated images, across all the five testing sets, the mean dice coefficients for the healthy/glaucoma OCT images were: $0.92 \pm 0.05/0.92 \pm 0.03$ for the RNFL + prelamina, $0.83 \pm .04/0.84 \pm 03$ for the RPE, $0.95 \pm 0.01/ 0.96 \pm 0.07$ for all other retina layers, and $0.90 \pm 0.03/ 0.91 \pm 0.05$ for the choroid. The mean sensitivities for the healthy/glaucoma OCT images were $0.92 \pm 0.01/0.92 \pm 0.03$ for the RNFL + prelamina, $0.87 \pm 0.04/0.88 \pm 0.03$ for the RPE, $0.96 \pm 0.04/0.96 \pm 0.03$ for all other retina layers, and $0.89 \pm 0.06/0.91 \pm 0.02$ for the choroid respectively. For all the tissues, the mean specificities were always above $0.99$ for both glaucoma and healthy subjects. In all experiments, there were no significant differences (mean dice coefficients, specificities and sensitivities) in the performance of digital staining between glaucoma and healthy OCT images (\textbf{Figure \ref{fig:5}}).

\begin{figure}[H]
    \centering
    \includegraphics[width=1.0\textwidth]{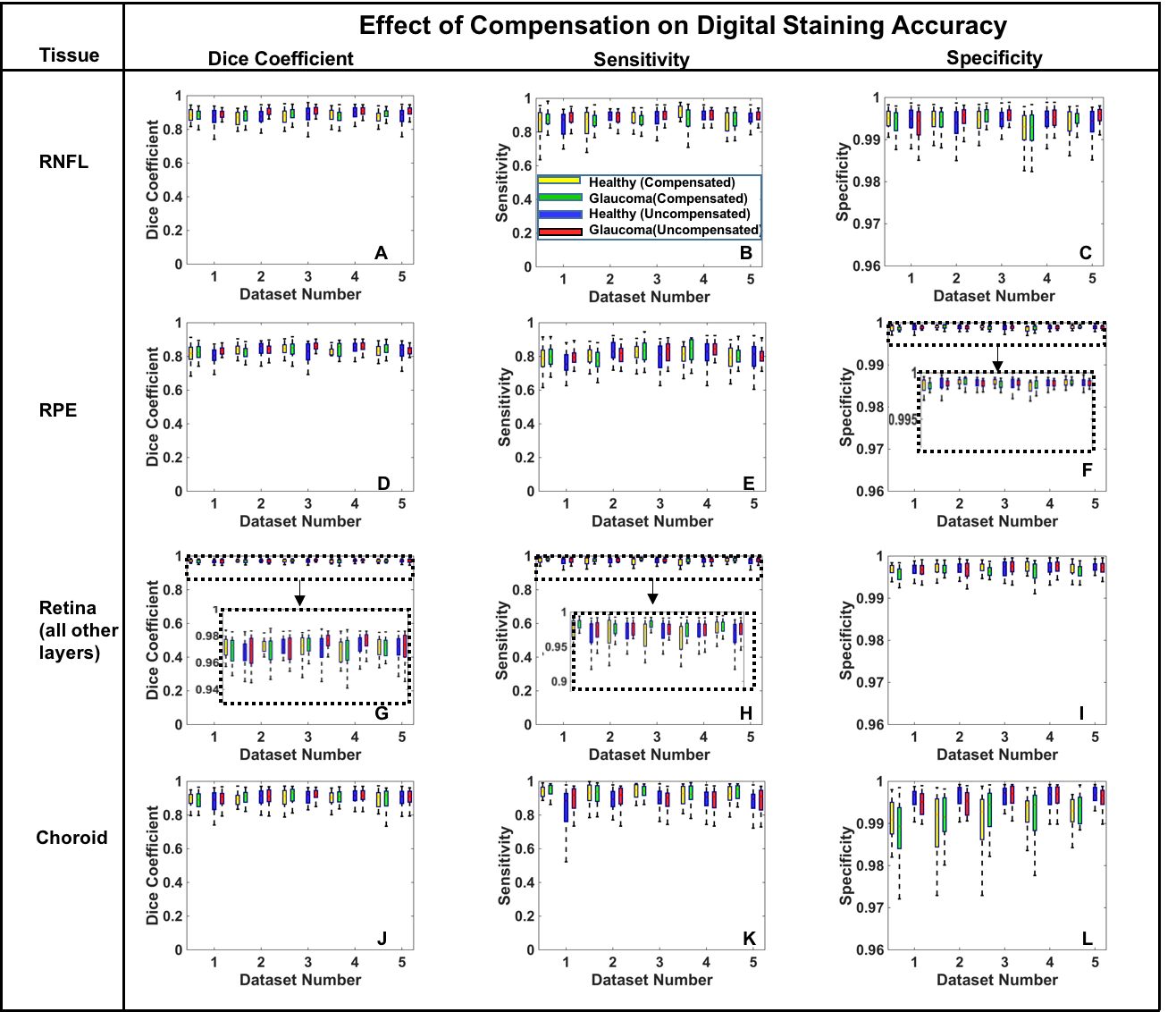}
    \caption{The effect of compensation on the digital staining accuracy is presented. A total of 5 compensated and uncompensated datasets were used for training (40 images) and its corresponding testing (60 images). (A-C) represent the dice coefficients, sensitivities and specificities as box plots for the RNFL + prelamina for compensated (healthy in yellow; glaucoma in green) and uncompensated images (healthy in blue; glaucoma in red). (D-F) represent the same for the RPE, (G-I) represent the same for all other retinal layers and (J-L) represent the same for the choroid.}
    \label{fig:6}
\end{figure}

Further, the performance of the digital staining did not significantly improve when using compensation (\textbf{Figure \ref{fig:6}}). 

\begin{figure}[H]
    \centering
    \includegraphics[width=1.20\textwidth]{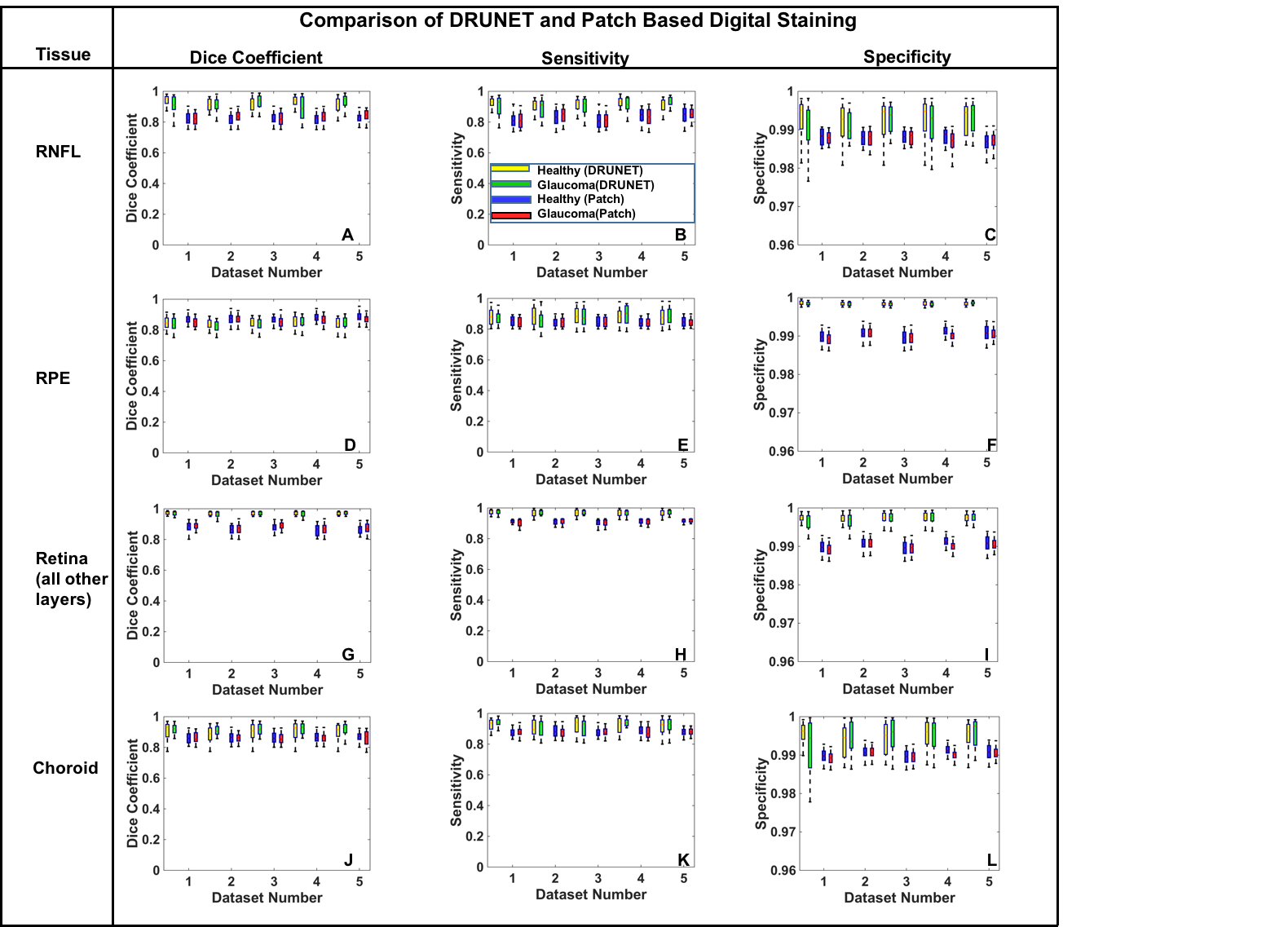}
    \caption{A comparative study between DRUNET and the patch-based digital staining approach is presented. A total of 5 datasets were used for training (40 images) the DRUNET/patch based architecture and its corresponding testing (60 images). (A-C) represent the dice coefficients, sensitivities and specificities as box plots for the RNFL + prelamina obtained from DRUNET (healthy in yellow; glaucoma in green) and the patch-based method (healthy in blue; glaucoma in red).  (D-F) represent the same for the RPE, (G-I) represent the same for all other retinal layers and (J-L) represent the same for the choroid.}
    \label{fig:7}
\end{figure}

Overall, the DRUNET performed significantly better for all the tissues compared to the patch-based approach, except for the RPE, in which case it performed similar (\textbf{Figure \ref{fig:7}}).

\section{Discussion}

In this study, we present DRUNET, a deep learning approach that is able to capture both local and contextual features to simultaneously stain (i.e. highlight) connective and neural tissues in OCT images of the ONH. The proposed study leverages on the inherent advantages of skip connections, residual learning and dilated convolutions. Having successfully trained, tested and validated on the OCT images from 100 subjects, we were able to consistently achieve a good qualitative and quantitative digital staining. Thus, we may be able to offer a robust segmentation framework, for the automated parametric study of the ONH tissues. \\

Using DRUNET, we were able to simultaneously isolate the RNFL+prelamina, the RPE, all other retinal layers, the choroid, the peripapillary sclera, the LC, noise and the vitreous humor with good accuracy. When trained and tested on compensated images, there was good agreement with manual segmentation, with the overall dice coefficient (mean of all tissues) being $0.91 \pm 0.04$ and $0.91 \pm 0.06$ for glaucoma and healthy subjects respectively. The mean sensitivities for all the tissues were $0.92 \pm 0.04$ and $0.92 \pm 0.04$ for glaucoma and healthy subjects respectively while the mean specificities were always higher than $0.99$ for all cases.\\

We observed that DRUNET offered no significant differences in the performance of digital staining when tested upon compensated (blood vessel shadows removed), or uncompensated images, as opposed to our previous patch-based method \cite{RN516}, that performed better on compensated images. This may be attributed to the extensive online data augmentation we used herein that also included occluding patches to mimic the presence of blood vessel shadows. In uncompensated images, the presence of retinal blood vessel shadows typically affects the automated segmentation of the RNFL \cite{RN527,RN522}, that can yield incorrect RNFL thickness measurements. This phenomenon may be more pronounced in glaucoma subjects that exhibit very thin RNFL. Our DRUNET framework, being invariant to the presence of blood vessel shadows, could potentially be extended to provide an accurate and reliable measurement of RNFL thickness. We believe this could improve the diagnosis and management of glaucoma. However, given the benefits of compensation in enhancing deep tissue visibility \cite{RN249}, and contrast \cite{RN20}, it may be advised to digitally stain compensated images for a reliable clinical interpretation of the isolated ONH tissues.\\

When trained and tested with the same cohort, DRUNET offered smooth and accurate delineation of tissue boundaries with reduced false predictions. Thus, it performed significantly better than the patch-based approach for all the tissues, except for the RPE, in which case it performed similarly. This may be attributed to DRUNET’s ability in capturing both local (tissue texture) and contextual features (spatial arrangement of tissues), compared to the patch-based approach that captured only the local features.\\

DRUNET consisted of 40,000 trainable parameters as opposed to the patch-based approach that required 140,000 parameters. Besides, DRUNET also eliminated the need for multiple convolutions on similar sets of pixels as seen in patch-based approach. Thus, DRUNET offers a computationally inexpensive and faster segmentation framework that only takes 80 ms to digitally stain one OCT image. This could be extended for the real time segmentation of OCT images as well. We are currently exploring such an approach.\\

We found that DRUNET was able to dissociate the LC from the peripapillary sclera. This provides an advantage as opposed to previous techniques that were able to segment only the LC \cite{RN536,RN326}, or the LC fused with the peripapillary sclera \cite{RN516}. To the best of our knowledge, no automated segmentation techniques have been proposed to simultaneously isolate all individual ONH connective tissues. We believe our network was able to achieve this because we used the Jaccard Index as part of the loss function. During training, by computing the Jaccard Index for each tissue, the network was able to learn the representative features equally for all tissues. This reduced the inherent bias in learning features of a tissue represented by a large number of pixels (e.g., retinal layers) as opposed to a tissue represented by a small number of pixels (e.g., LC/RPE). \\

We observed no significant differences in the performance of digital staining when tested on glaucoma or healthy images. The progression of glaucoma is characterized by thinning of the RNFL \cite{RN29,RN6,RN3} and decreased reflectivity (attenuation) of the RNFL axons \cite{RN526}, thus reducing the contrast of the RNFL boundaries. Existing automated segmentation tools for the RNFL rely on these boundaries for their segmentation and are often prone to segmentation artifacts \cite{RN518,RN519,RN520,RN521} (incorrect ILM/ posterior RNFL boundary), resulting in inaccurate RNFL measurements. This error increases with the thinning of the RNFL \cite{RN522}. Thus, glaucomatous pathology increases the likelihood of errors in the automated segmentation of the RNFL, leading to under- or over-estimated RNFL measurements that may affect the diagnosis of glaucoma \cite{RN522}. An automated segmentation tool that is invariant to the pathology is thus highly needed to robustly measure RNFL thickness. We believe DRUNET may be a solution for this problem, and we aim to test this hypothesis in future works. \\

With the proposed methodology, one could extract key structural parameters from the isolated tissues as an immediate clinical application of our technique. For instance, the peripapillary RNFL/choroidal thickness could be computed from digitally stained images as the number of vertical coloured pixels representing it, multiplied by a physical scaling factor. Upon isolation of the RPE from the central scan, the Bruch's membrane opening (BMO) points can be identified its end tips. The minimum-rim width can then be computed as the minimum distance between the BMO points and the inner limiting membrane \cite{RN543} (also obtained from digital staining). Further, one could also obtain key connective tissue parameters such as anterior LC insertion distance \cite{RN545}, LC surface depth \cite{RN546}, prelaminar thickness \cite{RN544} and prelaminar depth \cite{RN544} in a similar manner from the isolated sclera and LC. We believe, obtaining these parameters can improve the diagnosis, management and risk profiling of glaucoma. We are currently working on such a clinical translation of the proposed approach.\\

\noindent
There are several limitations to this study that warrant further discussion. 
\begin{itemize}
\item The accuracy of the algorithm was validated against the manual segmentations provided by a single expert observer (SD). The future scope of this study would be to provide a validation against multiple expert observers. Nevertheless, we offer a proof of concept for the simultaneous digital staining of the ONH tissues in OCT images.
\item The algorithm was trained with the images from a single machine (Spectralis). Currently, it is unknown if the algorithm would perform the same way if tested on images from multiple OCT devices. We are exploring other options to develop a device-independent digital staining approach.
\item We observed irregular LC boundaries that were inconsistent with that of the manual segmentations in few images. When extended for the automated parametric study, this could affect the LC parameters such as LC depth \cite{RN387}, LC curvature \cite{RN545}, and the global shape index \cite{RN376}. Given the significance of LC morphology in glaucoma \cite{RN396,RN387,RN376,RN397,RN401,RN403,RN395,RN394,RN228}, a more accurate delineation of the LC boundary would be required to obtain reliable parameters for a better understanding of glaucoma. This could be addressed using transfer learning \cite{RN531, RN529} by incorporating more information about LC morphology within the network. We are currently exploring such an approach.
\item A quantitative validation of the peripapillary sclera and the LC could not be performed as their true thickness could not be obtained from the manual segmentations due to limited visibility \cite{RN249}. 
\item We were unable to provide further validation to our algorithm by comparing it with data obtained from histology. This is a challenging task, given that one would need to image a human ONH with OCT, process it with histology and register both datasets. However, it is important to bear in mind that the understanding of OCT ONH anatomy stemmed from a single comparison of a normal monkey eye scanned in vivo at an IOP of 10 mm Hg and then perfusion fixed at time of sacrifice at the same IOP \cite{RN511}. Our algorithm produced tissue classification results that match the expected relationships obtained in this above-mentioned work. The absence of published experiments matching human ONH histology to OCT images, at the time of writing this paper, inhibits an absolute validation of our proposed methodology. 
\item A robust and accurate isolation of the ganglion cell complex (GCC) \cite{RN506} and the photoreceptor layers \cite{RN507}, whose structural changes are associated with the progression of glaucoma was not possible in both compensated and uncompensated images. The limitation of an accurate intra-retinal layer segmentation from ONH images can be attributed to the inherent speckle noise and intensity inhomogeneity \cite{RN517} which affects the robust delineation of the intra-retinal layers. This could be resolved by using advanced pre-processing techniques for image denoising (e.g. deep learning based) or a multi-stage tissue isolation approach (i.e., extraction of retinal layer followed by the isolation of intra-retinal layers).
\item Given the limitation of a small dataset (100 images), and the need for performing multiple experiments (repeatability), we were able to use only 40 images for training (60 images for testing) in each experiment. It is currently unknown if the performance of digital staining would improve when trained upon a larger dataset. Also, we would also like to emphasize again that there was no mixing of the training and testing sets in a given experiment. However, across all the experiments, there was indeed a small leakage of the testing/training sets. Nevertheless, we offer a proof of principle for a robust deep learning approach to digital stain ONH tissues that could be used by other groups for further validation. 
\end{itemize}

In conclusion, we have developed a deep learning algorithm for the simultaneous isolation of the connective and neural tissues in OCT images of the ONH. Given that the ONH tissues exhibit complex changes in their morphology with the progression of glaucoma, their simultaneous isolation may be of great interest for the clinical diagnosis and management of glaucoma.

\section*{Funding}

Singapore Ministry of Education Academic Research Funds Tier 1 (R-155-000-168-112 [AHT]); National University of Singapore (NUS) Young Investigator Award Grant (NUSYIA$\textunderscore$FY16$\textunderscore$P16; R-155-000-180-133; AHT); National University of Singapore Young Investigator Award Grant (NUSYIA$\textunderscore$FY13$\textunderscore$P03; R-397-000-174-133 [MJAG]); Singapore Ministry of Education Tier 2 (R-397-000-280-112 [MJAG]);National Medical Research Council (Grant NMRC/STAR/0023/2014 [TA]).\\

\section*{Disclosures}
The authors declare that there are no conflicts of interest related to this article.\\

\bibliographystyle{unsrt}
\bibliography{DRUNETarxiv}

\end{document}